\documentclass[runningheads,a4paper]{llncs}
\pdfoutput=1
\usepackage{times}
\usepackage{amssymb}
\usepackage{graphicx}
\usepackage{url}

\usepackage{xcolor}
\usepackage[most]{tcolorbox}

\newcommand{\keywords}[1]{\par\addvspace\baselineskip
\noindent\keywordname\enspace\ignorespaces#1}

\urldef{\mailsa}\path|{firstname.lastname, tav9580}@thi.de|

\begin{document}

\title{Measuring Sentiment Bias in Machine Translation}


\author{Kai Hartung\inst{1} \and Aaricia Herygers\inst{1} \and Shubham Kurlekar\inst{1} \and Khabbab Zakaria\inst{1} \and Taylan Volkan\inst{1} \and S\"{o}ren Gr\"{o}ttrup \inst{1} \and Munir Georges\inst{1,2}}


 \authorrunning{Kai Hartung et al.}

\institute{AImotion Bavaria, Technische Hochschule Ingolstadt, Germany \\
\mailsa\\
 \and
Intel Labs, Germany \\
}

\index{Hartung, Kai}
\index{Herygers, Aaricia}
\index{Kurlekar, Shubham}
\index{Zakaria, Khabbab}
\index{Volkan, Taylan}
\index{Gr\"{o}ttrup, S\"{o}ren}
\index{Georges, Munir}

\toctitle{} \tocauthor{}

\maketitle

%
%
%
%
\begin{abstract}

Biases induced to text by generative models have become an increasingly large topic in recent years.
In this paper we explore how machine translation might introduce a bias in sentiments as classified by sentiment analysis models.
For this, we compare three open access machine translation models for five different languages on two parallel corpora to test if the translation process causes a shift in sentiment classes recognized in the texts.
Though our statistic test indicate shifts in the label probability distributions, we find none that appears consistent enough to assume a bias induced by the translation process.

\end{abstract}

\section{Introduction
}\label{sec:intro}



With the increasing use of artificial intelligence also came a rise in research into its trustworthiness and fairness \cite{liu2022trustworthyAI}.
These studies have shown that models in various fields contain biases.
For instance, in computer vision a range of biases 
has been investigated, 
followed by novel mitigation techniques \cite{wang2020fairness-visual-recognition}. 
Similarly, biases have been found and mitigated \cite{zhang22n_interspeech} in speech recognition and recommender systems \cite{chen2023recommender-bias}.

In order to avoid perpetuating social biases and further contribute to discrimination, researchers have also studied biases in machine translation (MT).
For example, Prates et al. \cite{prates2020assessing} conducted a case study into Google Translate by translating sentences such as ``He/She is an engineer'', with varying occupations, from gender-neutral languages 
into English.
They found that, especially for job titles in the fields of science, technology, engineering, and mathematics, the English gendered pronoun tended to be male.
Following these results, Escudé et al. \cite{escude-font-costa-jussa-2019-equalizing} experimented with a debiasing method and a gender-neutral MT system \cite{zhao-etal-2018-learning}.
The study showed that a system that learned gender information disregarded contextual gender information
, providing wrong translations.
The gender-neutral system, however, did take the sentential gender information, providing correct translations, enabling it to achieve a higher BLEU \cite{papineni-etal-2002-bleu} performance.

Another type of bias is \textit{sentiment bias}, in which case a sentiment classification or sentiment analysis (SA) model provides a sentence with a sentiment (e.g., positive or negative) which may change when the phrasing or a certain word changes.
In \cite{huang-etal-2020-reducing}, a language generation model was used to generate sentences with varying occupations, countries, and gendered names.
Through counterfactual evaluation \cite{garg2019counterfactual}
, the study revealed that there were systematic differences in the sentiments across the varying inputs.
For example, sentences containing the word `baker' had a more positive sentiment than those containing the word `accountant'.

This paper thus seeks to measure sentiment bias in machine translation.
The following sections describe related work (Section~\ref{sec:lit}), followed by a description of our analysis (Section~\ref{sec:analysis}), the used models (Section~\ref{sec:models}) and corpora (Section~\ref{sec:corpora}).
The results are presented and discussed in Section~\ref{sec:results-disc}, followed by a conclusion in Section~\ref{sec:conc}.

\section{Related Work
}\label{sec:lit}

As MT is essentially a task on natural language processing in one language and generation in another, this section briefly describes related work on biases in natural language processing.
Extended literature overviews on such biases are provided in \cite{sun-etal-2019-mitigating,blodgett-etal-2020-language}.

In \cite{zhao-etal-2017-men} the authors studied language as support for visual recognition tasks.
Specifically, they investigated the data as well as the models for two tasks: multi-label object classification and visual semantic role labeling.
They found that the dataset contained a gender bias, which was amplified when used to train a model (e.g., `cooking' was 30\% more likely to be accompanied by `woman' than `man' in the training set, followed by a 70\% more likely association after training).
In order to counter this effect, the authors proposed to provide a balanced gender ratio for each of the activities, i.e., put constraints on the corpus level, within a framework called `Reducing Bias Amplification'.
This approach was able to decrease the bias by over 40\% for both studied tasks.

Furthermore, Zhao et al. \cite{zhao-etal-2018-gender-coreference} studied gender coreference (e.g., ``The physician called the secretary and told \textit{him/her} to cancel the appointment.'') on the WinoBias benchmark in a rule-based, a feature-rich, and a neural coreference system.
All three systems were found to stereotypically link occupations to gendered pronouns.
With existing word embedding debiasing techniques combined with data augmentation, the bias in WinoBias was removed.



Through a systematic study of text generated by two language models based on prompts mentioning various demographics, Sheng et al. \cite{sheng-etal-2019-woman} revealed that both models contained various biases.
The context for the biases was categorized as either `respect' (e.g., ``XYZ was known for'') or `occupation' (e.g., ``XYZ worked as'').
One model, GPT-2 \cite{Language-Models-are-Unsupervised-Multitask-Learners}, showed bias against black and gay people for both `respect' and `occupation', but was biased against men in the former and against women in the latter category.


A quantification, analysis, and mitigation of gender bias were also carried out for the contextualized word vectors of the ELMo \cite{peters-etal-2018-deep} model \cite{zhao-etal-2019-gender-word-embeddings}.
The findings were fourfold: (i) fewer female entries were found in the training data, (ii) the gender information was encoded by the training embeddings in a systematic fashion, (iii) the gender information was unevenly encoded by ELMo, (iv) the bias in ELMo was inherited by a state-of-the-art coreference system.
Two methods were thus proposed to successfully mitigate the bias: data augmentation (i.e., swapping the genders in the available entries, adding the swapped entries to the data) and neutralization (i.e., generating gender-swapped data).


Bordia and Bowman \cite{bordia-bowman-2019-identifying} proposed a metric to measure gender bias in both a corpus and a text generated by a model trained on that corpus, followed by a proposal for a regularization loss term. 
It was effective up until a certain weight, after which the model became unstable.
Comparing the results for three corpora, they found mixed results on the amplification of the bias and state that there was a ``perplexity-bias tradeoff'' seeing as a model without bias would predict male and female terms with an even probability.

Another approach to reduce gender stereotyping was proposed in \cite{zmigrod-etal-2019-counterfactual}, i.e., augmenting the dataset with counterfactual data (e.g., `el ingeniero' would become *`el ingeniera') and implementing a Markov Random Field for the agreement on a morpho-syntactic level.
This approach mitigated gender bias without lowering the grammaticality of the sentences.

Furthermore, Jia et al. \cite{jia-etal-2020-mitigating} investigated the amplification of gender bias through the lens of distribution.
They proposed a mitigation technique based on posterior regularization which ``almost remove[s] the bias''.



\section{Sentiment Bias Analysis}\label{sec:analysis}

To investigate whether the MT procedure changes expressions, we first do the translations and then perform SA on the back-translated version of the text.
We thus first take the text in its original language $c_{l_1}$ and translate it into an intermediary language $l_2$ to get $t_{l_2}(c_{l_1})$.
This translation we then translate back to $l_1$ to get $t_{l_1}(t_{l_2}(c_{l_1}))$.
For further comparison, we also translate the original version of the text $c_{l_2}$ in language $l_2$ into $l_1$ to get a second machine translated text $t_{l_1}(c_{l_2})$.
This allows us to compare the influence of the back-translation $l_2 \to l_1$ with that of the first translation $l_1 \to l_2$, which should only be visible in $t_{l_1}(t_{l_2}(c_{l_1}))$, but not in $t_{l_1}(c_{l_2})$.

We then apply the analyses on the three versions of the same text in the same language $c_{l_1}$, $t_{l_1}(t_{l_2}(c_{l_1}))$,  $t_{l_1}(c_{l_2})$.
To compare the SA results we look at three metrics.
A previously used metric in the context of bias analysis is the Wasserstein distance \cite{huang-etal-2020-reducing,jiang2020wasserstein} (WD) between two distributions.
In this case, we compare those over the probability scores assigned to each sentiment label.
This distance measure however has no set scale or threshold that allows for making a statement about whether two distributions are distant enough to constitute a bias.
To get a better impression whether an actual bias is present, we also perform two statistical tests: the paired t-test applied again on the probability scores of the class labels and the $\chi^2$-test on the class labels themselves.
For both tests, the null hypothesis assumes that the distributions are equal, so in cases where we reject the null hypothesis the distributions are unequal and we can assume the translation had a notable effect on the sentiment classes.

\section{Models}\label{sec:models}

In this section, we first introduce the translation models before the sentiment analysis models used in our studies are described.

\subsection{Translation models}\label{ssec:translation-models}

We used three MT tools which offer pre-trained models.
The languages we addressed are German (de), English (en), Spanish (es), Hebrew (he), and Chinese (zh).


No Language Left Behind (\textit{fairseq-nllb}) \cite{No-Language-Left-Behind:-Scaling-Human-Centered-Machine-Translation} is an attempt to address data scarcity for translation models for low-resource languages, resulting in a multilingual model able to translate between 204 languages.
%
%
%
The authors used a large-scale mining approach to create a dataset of over 1.1 billion sentence pairs.
Additionally, they created the NLLB-SEED dataset, comprising human-translated lines from Wikipedia for 39 languages.
For evaluation, the FLORES-200 dataset was created, comprising 3001 English samples from web articles translated into 204 languages by human experts.
The translation model itself is based on the Encoder-Decoder Transformer architecture proposed by \cite{Attention-is-All-You-Need}.
However, the authors used pre-layer-normalization for each transformer sub-layer instead of applying layer-normalization after the residual connections.
In addition, the model was built as a Sparsely Gated Mixture of Experts \cite{almahairi-dynamic-2016,bengio-estimating-2013,shazeer-utrageouslylarge-2017o,lepikhin-gshard-2020}.
For every fourth transformer block the fully connected layer was split into a number of experts, each consisting of a separate fully connected layer followed by a softmax layer.
Each token is then assigned to the top 2 experts according to the Top-k-Gating algorithm \cite{lepikhin-gshard-2020}.
The appropriate routing of the tokens is optimized through an additional training loss.
This architecture enables training on several translation directions at once, without much cross-lingual transfer interference on low-resource languages.
During training a curriculum learning strategy is applied, in which all language pairs are divided into buckets.
Each bucket is introduced after a number of updates, based on the median number of updates, after which all directions in that bucket would start to overfit.


Argos-translate \cite{Finlay_Argos_Translate} is an open source offline translation library.
It uses the OpenNMT \cite{klein-etal-2017-opennmt} sequence-to-sequence transformer to train specific language pair models.
For language pairs that do not have a direct translation model between them, an intermediate language such as English is used to accomplish the task.
This allows the library to translate between a wide variety of languages at the cost of some loss in translation quality.
The training data is collected from OPUS \cite{Parallel-Data-Tools-and-Interfaces-in-OPUS}.
Wiktionary \cite{Wiktextract-Wiktionary-as-Machine-Readable-Structured-Data} definition data is used to improve the translation quality of low-resource languages and single-word translations.
The architecture of OpenNMT is based on sequence-to-sequence learning with attention based on \cite{Sequence-to-Sequence-Learning-with-Neural-Networks} and rewritten for ease of efficiency and readability.
The sequence-to-sequence model uses subword units, given by SentencePiece \cite{SentencePiece-A-simple-and-language-independent-subword-tokenizer-and-detokenizer-for-Neural-Text-Processing}.

The BERT2BERT encoder-decoder model was introduced by Rothe et al. \cite{Leveraging-Pre-trained-Checkpoints-for-Sequence-Generation-Tasks}.
The authors developed a transformer-based sequence-to-sequence model compatible with the publicly available pre-trained checkpoints of commonly-used models: BERT \cite{BERT:-Pre-training-of-Deep-Bidirectional-Transformers-for-Language-Understanding}, GPT-2 \cite{Language-Models-are-Unsupervised-Multitask-Learners}, and RoBERTa \cite{RoBERTa:-A-Robustly-Optimized-BERT-Pretraining-Approach}.
BERT (Bidirectional Encoder Representations from Transformers) was proposed in order to improve the fine-tuning based approaches.
For BERT2BERT, where both the encoder and decoder are BERT, there are 221M trainable parameters. Of them 23M are embedding parameters, 195M parameters are initialized from checkpoint and 26M parameters are initialized randomly.
The checkpoint was pre-trained on 108 languages using a multilingual Wikipedia dump with 110,000 words.

\subsection{Sentiment Analysis Models}\label{ssec:analysis-models}

We used SA tools for five languages:
GermanSentiment \cite{Training-a-Broad-Coverage-German-Sentiment-Classification-Model-for-Dialog-Systems}, German; Vader \cite{Vader:-A-parsimonious-rule-based-model-for-sentiment-analysis-of-social-media-text}, English; PySentimiento \cite{pysentimiento:-A-Python-Toolkit-for-Sentiment-Analysis-and-SocialNLP-tasks}, Spanish; HeBERT \cite{HeBERT-HebEMO:-a-Hebrew-BERT-Model-and-a-Tool-for-Polarity-Analysis-and-Emotion-Recognition}, Hebrew; and ASBA \cite{A-Modularized-Framework-for-Reproducible-Aspect-based-Sentiment-Analysis}, Chinese.

GermanSentiment \cite{Training-a-Broad-Coverage-German-Sentiment-Classification-Model-for-Dialog-Systems} is a SA model trained with the use for chatbot dialogue in mind, to better manage user feedback.
It has been trained on data from social media, review texts and service robot  field tests as well as additional neutral data from the Leipzig Corpora Collection \cite{Building-and-Using-Comparable-Corpora}.
The model itself is based on the BERT architecture \cite{BERT:-Pre-training-of-Deep-Bidirectional-Transformers-for-Language-Understanding}.
It predicts one of three classes for a sentence: positive, negative, neutral.
The authors report a Macro-F-score of up to 0.97 on their training data and 0.8 on an additional dataset not used during training.

Vader \cite{Vader:-A-parsimonious-rule-based-model-for-sentiment-analysis-of-social-media-text} is a rule based SA model for English.
It utilizes sentiment lexica for determining sentiment polarity and heuristics for determining intensity.
In addition to the three common classes positive, negative and neutral, the model can also label compound statements, which contain more than one sentiment.
Vader achieves an F1-score of  0.96 on social media text and 0.55 on New York Time Editorials.

PySentimiento \cite{pysentimiento:-A-Python-Toolkit-for-Sentiment-Analysis-and-SocialNLP-tasks} is a transformer-based model with support for Spanish and English.
The Spanish model is based on RoBERTuito \cite{RoBERTuito:-a-pre-trained-language-model-for-social-media-text-in-Spanish}, which follows the Roberta \cite{RoBERTa:-A-Robustly-Optimized-BERT-Pretraining-Approach} architecture and is trained on Spanish tweets.
The SA task for Spanish was trained on the TASS2020 Dataset \cite{Overview-of-TASS-2020:-Introducing-emotion-detection} which contains annotated tweets.
The labels are again positive, negative and neutral.
For this model the authors report a 0.7 Macro-F1 score.

HeBERT \cite{HeBERT-HebEMO:-a-Hebrew-BERT-Model-and-a-Tool-for-Polarity-Analysis-and-Emotion-Recognition} is a model based on the BERT architecture \cite{BERT:-Pre-training-of-Deep-Bidirectional-Transformers-for-Language-Understanding} finetuned for sentiment and emotion classification tasks.
The pretraining was done on Wikipedia and other web-based data and the SA was trained on crowd annotated comments on news articles.
As the above described models, HeBERT too classifies Sentiment in one of the three categories positive, negative, and neutral.
From their tests, the authors report 0.94 accuracy.

ASBA \cite{A-Modularized-Framework-for-Reproducible-Aspect-based-Sentiment-Analysis} is a framework for aspect-based SA offering models for several languages.
The Chinese model used here is based on a Chinese BERT \cite{BERT:-Pre-training-of-Deep-Bidirectional-Transformers-for-Language-Understanding} model and was trained on Chinese Opinion Analysis Data in the domains Phone, Camera, Notebook, Car in addition to MOOC data.
The reported performance of this model is an accuracy of 0.96 and F1 of 0.95.
This model only predicts two labels: positive and negative.

\section{Corpora}\label{sec:corpora}

\begin{table}[t!]
\centering
\begin{tabular}{l|cccccccccc}
  & de-en & de-es & de-he & de-zh & en-es & en-he & en-zh & es-he & es-zh & he-zh \\
 \hline
 TED2020 & 296K & 294K & 230K & 15K & 417K & 352K & 16K & 350K & 16K & 16K\\
Global Voices & 74K & 70K & 475 & 14K & 381K & 1K & 134K & 977 & 91K & 127 \\
\end{tabular}
\caption{Dataset sizes. For each dataset the number of lines for each language pair.}
\label{tab:dataset-sizes}
\end{table}

The corpora used in this work were taken from the open-source collection of parallel corpora OPUS \cite{Parallel-Data-Tools-and-Interfaces-in-OPUS}, which provides a compilation of aligned lines for each available corpus language pair.
TED2020 \cite{reimers-gurevych-2020-multilingual-sentence-bert} contains a crawled collection of nearly 4000 TED and TED-X transcripts dated July 2020.
A global community of volunteers translated the transcripts to 100 languages.
The corpus was created for the purpose of training multilingual sentence embeddings through knowledge distillation \cite{reimers-gurevych-2020-multilingual-sentence-bert} and has been used for domain-specific MT \cite{Results-of-the-WMT21-Metrics-Shared-Task:-Evaluating-Metrics-with-Expert-based-Human-Evaluations-on-TED-and-News-Domain,Building-Machine-Translation-System-for-Software-Product-Descriptions-Using-Domain-specific-Sub-corpora-Extraction}.
The Global Voices corpus \cite{Parallel-Data-Tools-and-Interfaces-in-OPUS} (version 2018q4) contains stories from the news website Global Voices\footnote{https://globalvoices.org/} and is also used in MT tasks such as domain-specific MT \cite{Building-Machine-Translation-System-for-Software-Product-Descriptions-Using-Domain-specific-Sub-corpora-Extraction}, data augmentation \cite{Simulated-multiple-reference-training-improves-low-resource-machine-translation}, or as part of low resource datasets \cite{The-FLoRes-Evaluation-Datasets-for-Low-Resource-Machine-Translation:-Nepali-English-and-Sinhala-English}.
The dataset sizes for both corpora are available in Table~\ref{tab:dataset-sizes}.

\section{Results and Discussion}\label{sec:results-disc}

\begin{figure}[t] 
   \centering
   \includegraphics[width=1\linewidth]{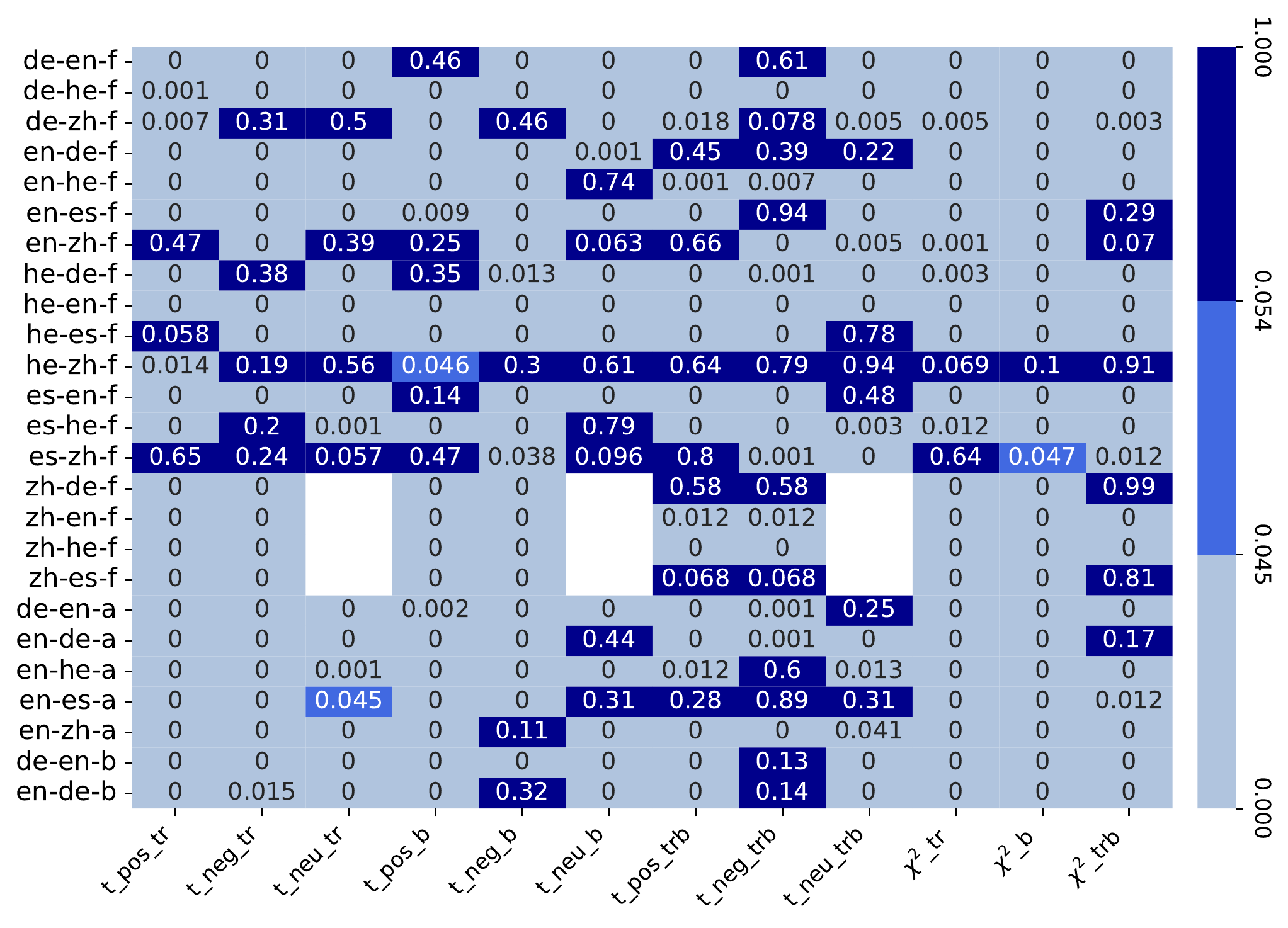}
   \caption{TED2020: p-values for each label distribution and language pair.
   ``\_tr'' in the x-axis compares the original $c_{l_1}$ to the translation $t_{l_1}(c_{l_2})$, while ``\_b'' compares $c_{l_1}$ to the back-translation $t_{l_1}(t_{l_2}(c_{l_1}))$ and ``\_trb'' compares translation with back-translation.
   The last letter in the y-axis describes the model; f: fairseq, a: Argos, and b: BERT2BERT.
   Equality is rejected below 0.05.}
   \label{fig:pval_TED}
 \end{figure}

 \begin{figure}[th] 
   \centering
   \includegraphics[width=1\linewidth]{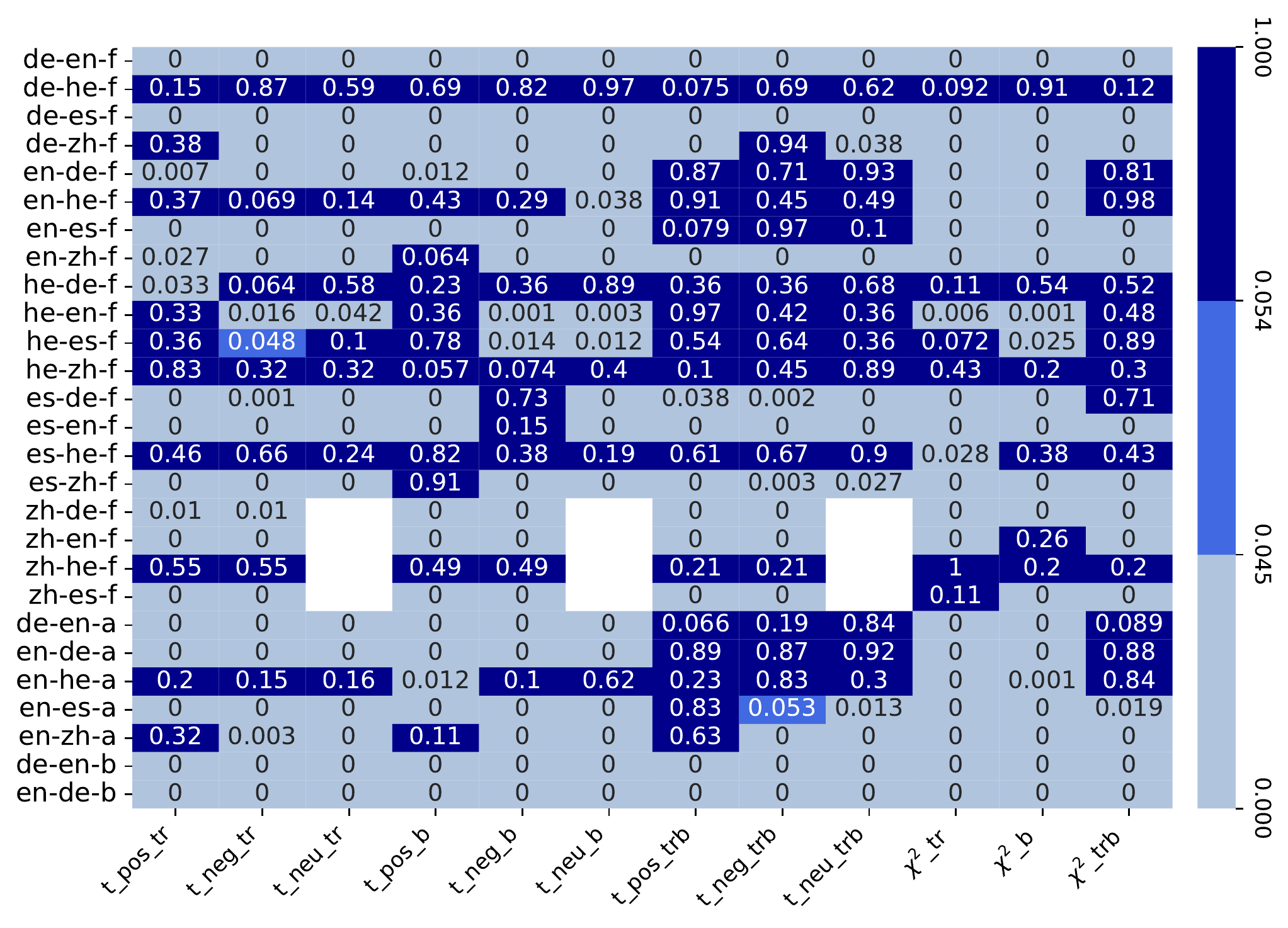}
   \caption{GlobalVoices: p-values for each label distribution and language pair.}
   \label{fig:pval_GV}
 \end{figure}

 \begin{figure}[th]
   \centering
   \includegraphics[width=1\linewidth]{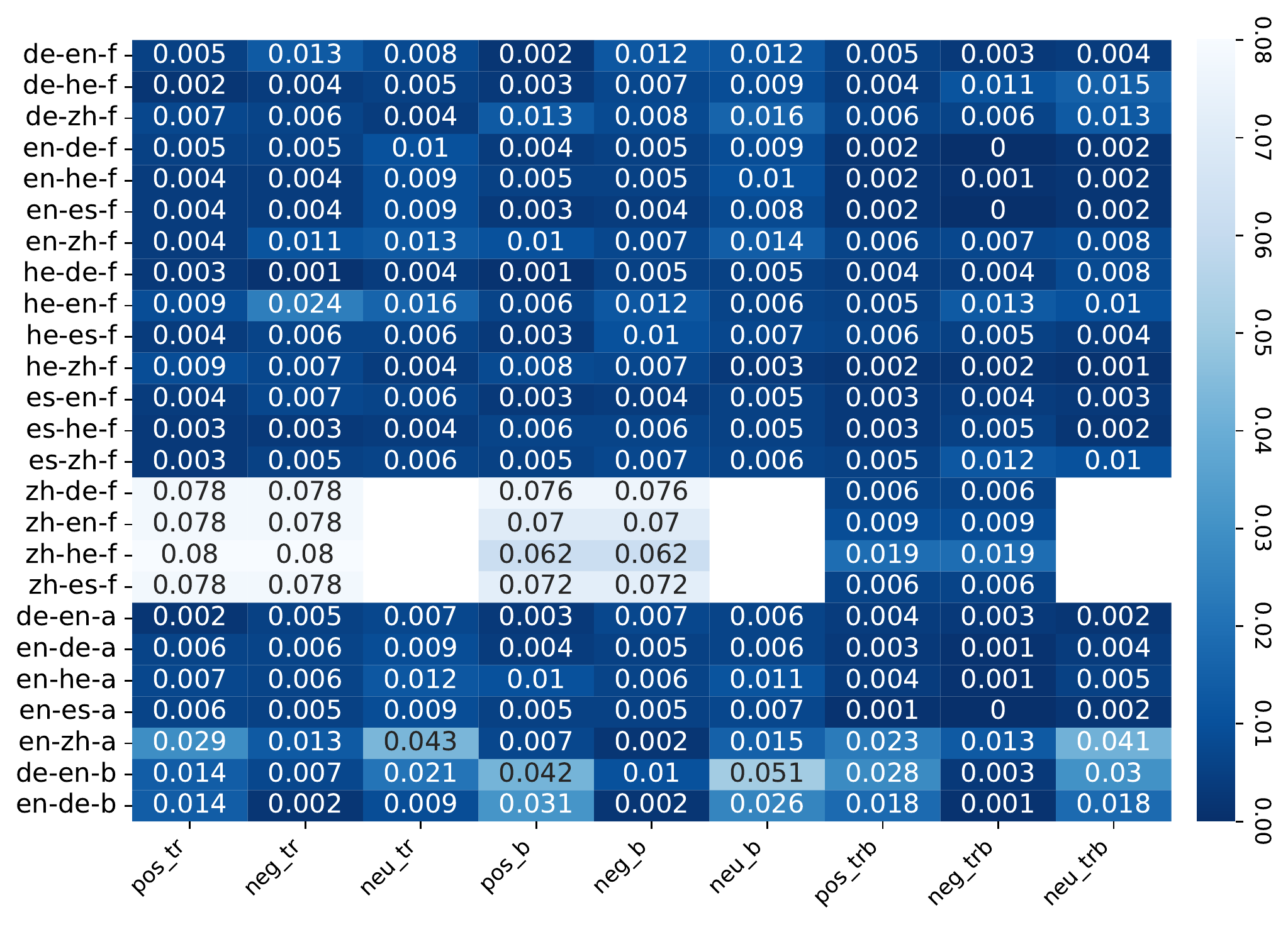}
   \caption{TED2020: Wasserstein distances for each label distribution and language pair.}
   \label{fig:wd_TED}
 \end{figure}

The heat maps in Figures~\ref{fig:pval_TED} and \ref{fig:pval_GV} show the results on the statistical tests for the TED2020 and Global Voices corpora respectively.
As a p-value below 0.05 for both t-test and $\chi^2$-test implies that equality of the compared distributions can be rejected, most translations appear to induce a shift in the distributions.
It is notable, though, that the smaller Global Voices corpus overall and the smaller sets including Hebrew, specifically, are the least likely to reject equality.

Furthermore the t-test for the comparison between translation $t_{l_1}(c_{l_2})$ and back-translation $t_{l_1}(t_{l_2}(c_{l_1}))$ in Global Voices also appear less likely to reject equality.
This might be due to the fact that both texts for these comparisons have been produced by the same translation model and are therefore more likely to share a specific vocabulary and structural style with each other than with the original text $(c_{l_1})$.
This in turn would lead to both translation to more likely share the same sentiment indicators to be recognized by the sentiment classifiers.

\begin{figure}[h!]
   \centering
   \includegraphics[width=1\linewidth]{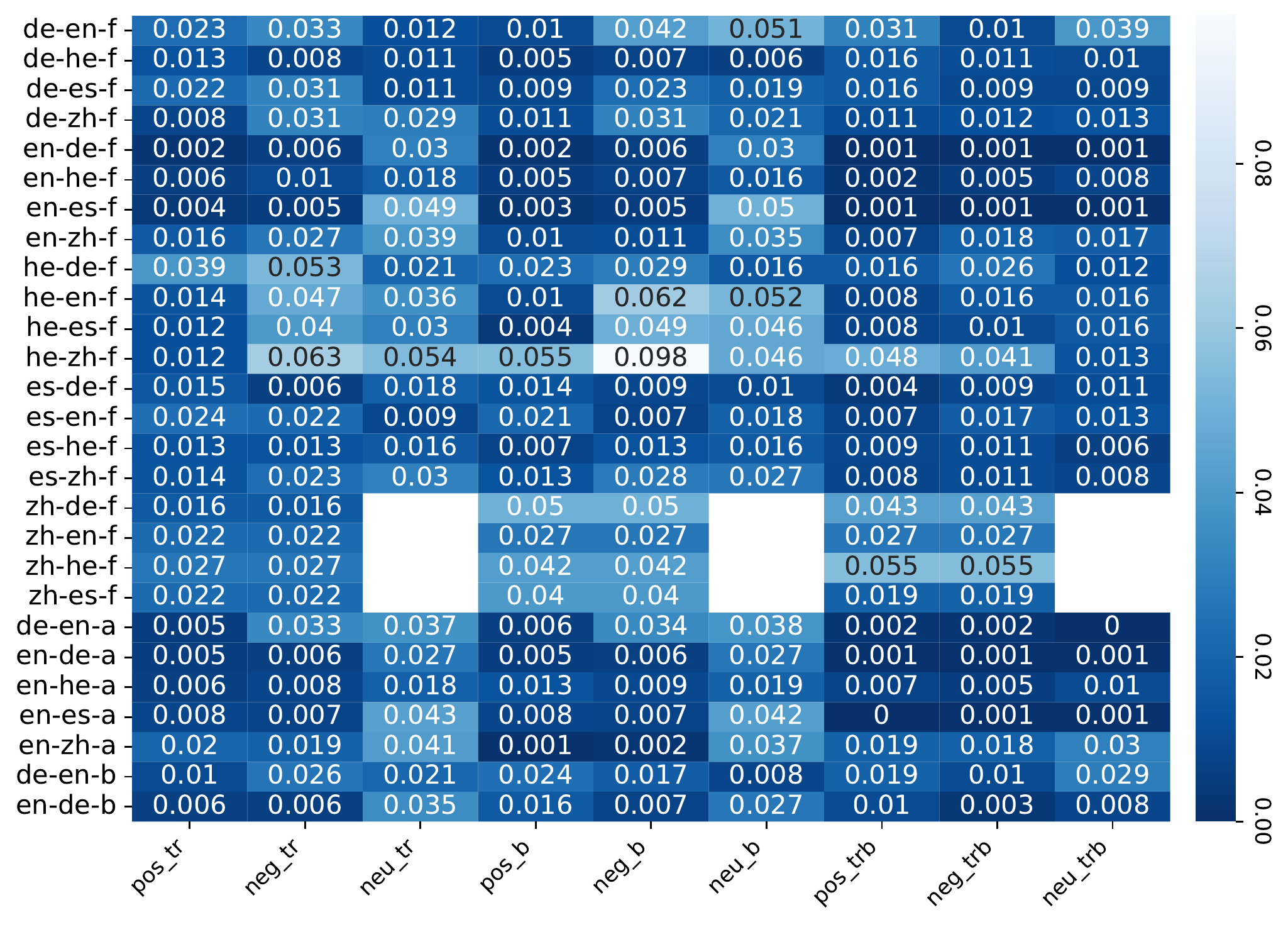}
   \caption{Global Voices: Wasserstein distances for each label distribution and language pair.}
   \label{fig:wd_GV}
\end{figure}

The same observations, however, can not be made from the WDs in Figures \ref{fig:wd_TED} and \ref{fig:wd_GV}.
Here the distances in Global Voices are overall larger and the Hebrew involving pairs do not stand out as having a particularly small distance.
On the other hand, the pairs classified in Chinese do stand out as having larger distances especially in TED2020.

To further test these observations, we compute the correlations between p-values and WDs per corpus and overall.
Linear regression for WD and p-value from t-test and $\chi^2$ return a correlation of -0.02 and -0.01 respectively.
This confirms that there is little relation between the distance and p-values.

Next, we test how the WD relates to the translation quality as measured by the BLEU score.
This comparison too shows no influence of the translation quality on the similarity of the sentiment scores with a correlation value of 0.

As the results from the statistic tests show shifts in the label probability distributions in several cases, we need to test for the direction of potential biases.
Thus, we 
apply one-directional t-tests in both directions and filter the results such that only the cases with the highest certainty are considered.
This means that we filter out cases where all instances shift in the same direction for all labels, cases where both directions reject equality, and cases where only a shift in one direction is observed over all three labels.

\begin{figure}[h!]
\centering
   \includegraphics[width=0.49\linewidth]{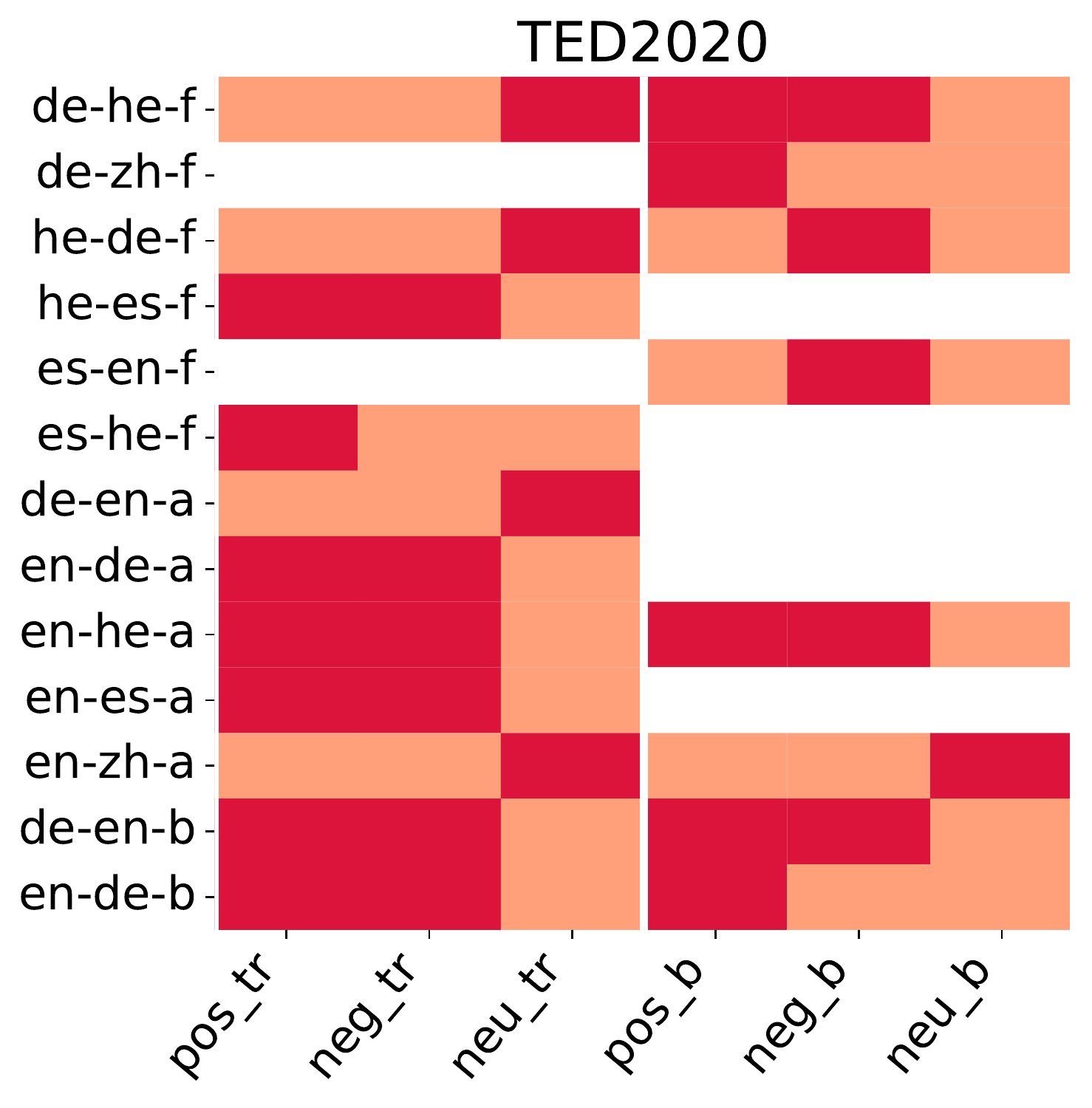}
   \includegraphics[width=0.49\linewidth]{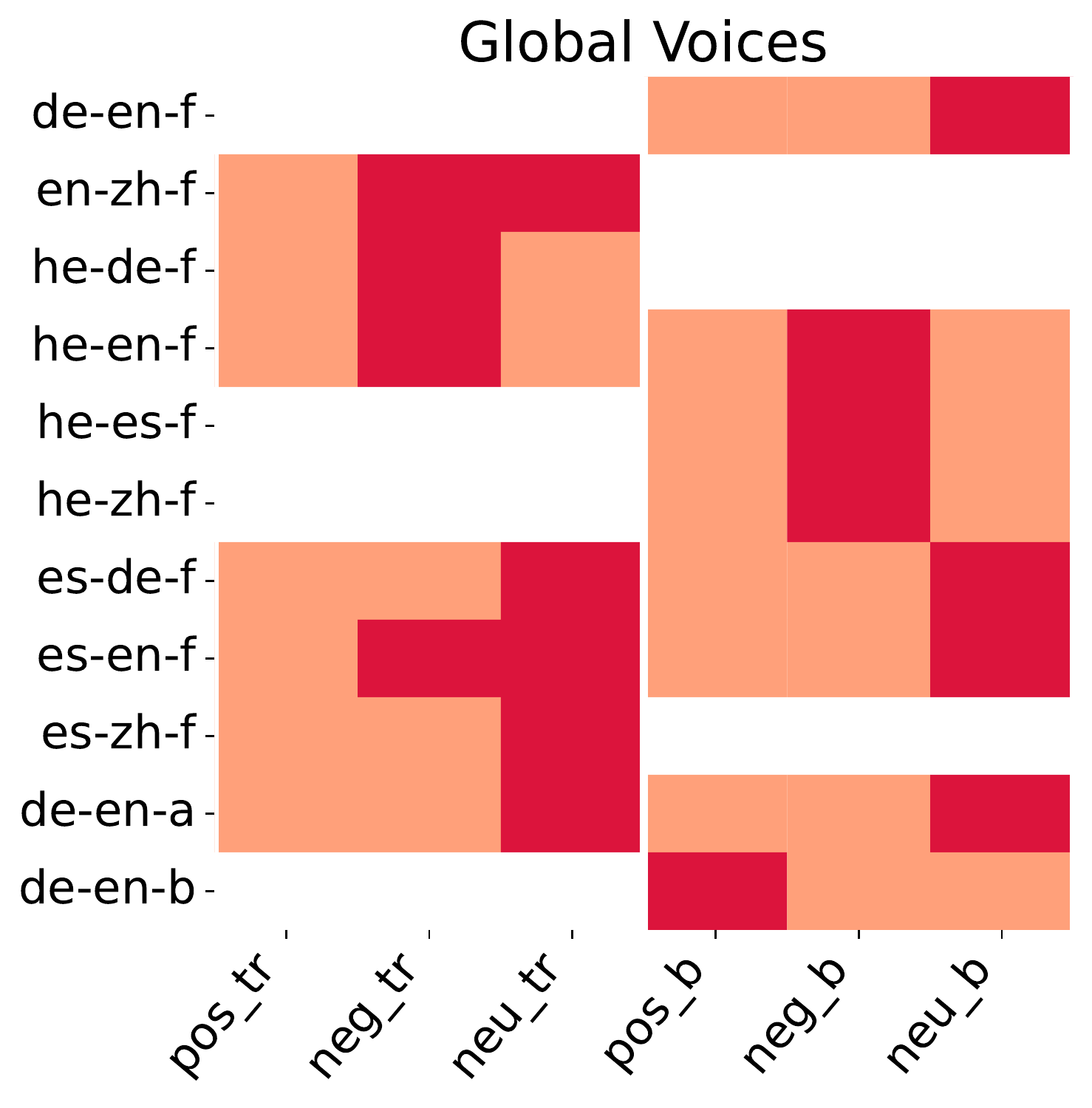}
   \caption{Shifts in label probability distributions.
   Red implies a shift occurred for that language pair, dark red indicates the label(s) towards which the translation caused a shift for direct translation (\_tr) or translation and back-translation (\_b).
   The y-labels describe the language pairs $l_1$-$l_2$ and the translation model: -f: fairseq, -a: argos, -b: BERT2BERT.
   }
   \label{fig:shift}
\end{figure}

The resulting cases with the highest certainty that the translation process caused a shift in class probability from one set of labels towards another are presented in Figure \ref{fig:shift}.
In most cases, the shifts from translation $t_{l_1}(c_{l_2})$ and back-translation $t_{l_1}(t_{l_2}(c_{l_1}))$ do not overlap.
Similarly, the shifts in translation for TED2020 and Global Voices mostly do not overlap either.
For example, in the TED2020Corpus, the German to Hebrew translation by fairseq (de-he-f) has a shift towards the neutral label after the translation and a respective reverse shift towards the positive and negative labels after the back-translation.
But for the same pair de-he-f no shifts at all can be found in the Global Voices corpus.

The one exception where both corpora, and for Global Voices both $t_{l_1}(c_{l_2})$ and $t_{l_1}(t_{l_2}(c_{l_1}))$, agree is the translation from German to English by the Argos system.
For TED2020 this also fits with the reverse shift for Argos' translation from English to German.
For the rest of the translations, no visible pattern across translation direction and corpus is apparent.

The WDs for this case are on the lower end of the spectrum for TED2020 and close to the mean in Global Voices.
Thus the WD cannot confirm a clear bias in the German-English pair.
The largest distances can be observed for the cases with Chinese as $l_1$, especially in TED2020.
These, however, cannot be confirmed as being consistently directed by the t-test.
In accordance to the overall low correlation between p-values and WD, the approaches do not agree on the seemingly most biased cases.

It is also worth pointing out, that the translations to Chinese achieve the worst BLEU-scores.
Pairs with Chinese as $l_1$ achieve an average score of 0.27 versus the average score over all other languages is 29.27.
So this might be a cause for the larger distances for these languages without a clearly directed shift.
Another possible cause for the higher distances for Chinese might lie in the fact that the SA model used for Chinese is the only binary classifier among those used in this study.
Therefore any random change from one label has only one option to shift towards and the sum of the random changes might appear as a shift in one direction without being actual bias.

\section{Conclusion}\label{sec:conc}

This study set out to explore whether MT systems introduce biases in sentiment expressions.
We compared three translation models (fairseq-nllb \cite{No-Language-Left-Behind:-Scaling-Human-Centered-Machine-Translation}, Argos-translate \cite{Finlay_Argos_Translate}, and BERT2BERT \cite{Leveraging-Pre-trained-Checkpoints-for-Sequence-Generation-Tasks}) for five languages (German, English, Hebrew, Spanish, and Chinese) from the TED2020 and Global Voices corpora.
Our statistical analyses (paired t-test and $\chi^2$-test) were not able to confirm any bias.
The closest to this is the translation from German to English by the Argo translation system, which causes a shift towards neutral sentiments for both corpora.
This `bias', however, cannot be substantiated by a notably large WD.

Future work might consider inspecting the differences between the target labels defined for different SA approaches, which may influence how label distributions shift through the translation process.
Another avenue to explore may be the effect of translations on more explicitly linguistic attributes of text, such as syntactic structure vocabulary or complexity.




\bibliographystyle{splncs04}
\bibliography{paper}

\end{document}